\documentclass[11pt]{article} 
\usepackage{amsfonts, amssymb, amsmath}
\usepackage{algorithm, algorithmic, amsthm}
\usepackage{graphicx}
\usepackage{color}
\usepackage{hyperref}
\usepackage[top=2.5cm, bottom=3cm, left=3cm, right=3cm]{geometry}

\usepackage{subfigure}

\usepackage{hyperref}
\usepackage{url}
\usepackage{graphicx, wrapfig}
\usepackage{amsfonts, amssymb, amsmath}
\usepackage{algorithm, algorithmic, amsthm,bm}
\usepackage{subfigure}
\usepackage{multirow}
\usepackage{soul}
\usepackage{enumitem}
\usepackage{color}
\usepackage{booktabs}
\usepackage{CJK}

{%
\begin{list}{#1}{
\vspace{-\topsep}
\vspace{-\partopsep}
\setlength{\itemindent}{0cm}
\setlength{\rightmargin}{0cm}
\setlength{\listparindent}{0cm}
\settowidth{\labelwidth}{#1}
\setlength{\leftmargin}{\labelwidth}
\addtolength{\leftmargin}{\labelsep}
\setlength{\itemsep}{0cm}
}%
}%
{%
\end{list}
\vspace{-\topsep}
\vspace{-\partopsep}
}

{\begin{enumerate}%
}%
{\end{enumerate}}

\theoremstyle{plain}

\newtheorem*{lemma*}{Lemma}

\newtheorem*{prop*}{Proposition}

\theoremstyle{definition}

\newtheorem*{defn*}{Definition}

\newtheorem*{exmp*}{Example}

\newtheorem*{conj*}{Conjecture}

\theoremstyle{remark}

\newtheorem*{rmk*}{Remark}

\title{Backward and Forward Language Modeling for Constrained Sentence Generation}
\author{\sf Lili Mou,$^1$ Rui Yan,$^2$ Ge Li,$^1$ Lu zhang,$^1$ Zhi Jin$^1$\\
$^1$Software Institute, Peking University\\
Key Laboratory of High Confidence Software Technologies (Peking University),\\
Ministry of Education, China\\
$^2$Baidu Inc.\\
{\tt\{doublepower.mou,rui.yan.peking\}@gmail.com}\\
{\tt\{lige,zhanglu,zhijin\}@sei.pku.edu.cn}
}
\date{}

\begin{document}
\begin{CJK}{UTF8}{gbsn}
\maketitle
\begin{abstract}
Recent language models, especially those based on recurrent neural networks (RNNs), make it possible to generate natural language from a learned probability. Language generation has wide applications including machine translation, summarization, question answering, conversation systems, etc. Existing methods typically learn a joint probability of words conditioned on additional information, which is (either statically or dynamically) fed to RNN's hidden layer. In many applications, we are likely to impose hard constraints on the generated texts, i.e., a particular word must appear in the sentence. Unfortunately, existing approaches could not solve this problem. In this paper, we propose a novel backward and forward language model. Provided a specific word, we use RNNs to generate previous words and future words, either simultaneously or asynchronously, resulting in two model variants. In this way, the given word could appear at any position in the sentence. Experimental results show that the generated texts are comparable to sequential LMs in quality. 
\end{abstract}

\section{Introduction}\label{sec:introduction}%
Language modeling  is aimed at minimizing the joint probability of a corpus.
It has long been the core of natural language processing (NLP) \cite{NLP}, and has inspired a variety of other models, e.g., the $n$-gram model, smoothing techniques \cite{smoothing}, as well as various neural networks for NLP \cite{NPLM,unified,SDP-LSTM}. In particular, the renewed prosperity of neural models has made groundbreaking 
improvement in many tasks, including language modeling \textit{per se} \cite{NPLM}, 
part-of-speech tagging, named entity recognition, semantic role labeling \cite{unified}, etc. 

The recurrent neural network (RNN) is a prevailing class of language models;
it is suitable for modeling time-series data (e.g., a sequence of words)
by its iterative nature. An RNN usually keeps one or a few hidden layers;
at each time slot, it reads a word, and changes its state accordingly.
Compared with traditional $n$-gram models, RNNs are more capable of learning long range features---especially with long short term memory (LSTM) units \cite{LSTM} or gated recurrent units (GRU) \cite{GRU}---and hence are better at capturing the nature of sentences.  On such a basis, it is even possible to generate a sentence from an RNN language model, which has wide applications in NLP, including machine translation \cite{seq2seq}, abstractive summarization \cite{summarization}, question answering \cite{QA}, and conversation systems \cite{AwI}. The sentence generation process is typically accomplished by choosing the most likely word 
at a time, conditioned on previous words as well as additional information depending on the task (e.g., the vector representation of the source sentence in a machine translation system \cite{seq2seq}).

In many scenarios, however, we are likely to impose constraints on the generated sentences. For example, a question answering system may involve analyzing the question and querying an existing knowledge base, to the point of which, a candidate answer is at hand. A natural language generator is then supposed to generate a sentence, coherent in semantics, containing the candidate answer. Unfortunately, using existing language models to generate a sentence with a given word is non-trivial: adding additional information \cite{sigdial,QA} about a word does not guarantee that the wanted word will appear; generic probabilistic samplers (e.g., Markov chain Monte Carlo methods) hardly applies to RNN language models\footnote{With recent efforts in \cite{biRNN}.}; setting an arbitrary word to be the wanted word damages the fluency of a sentence; imposing the constraint on the first word restricts the form of generated sentences. 

In this paper, we propose a novel backward and forward (B/F) language model to tackle the problem of constrained natural language generation. Rather than generate a sentence from the first word to the last in sequence as in traditional models, we use RNNs to generate previous and subsequent words conditioned on the given word. The forward and backward generation can be  accomplished either simultaneously or asynchronously, resulting in two variants, syn-B/F and asyn-B/F. In this way, our model is complete in theory for generating a sentence with a wanted word, which can appear at an arbitrary position in the sentence.

The rest of this paper is organized as follows. Section \ref{sec:Related} reviews existing language models and natural language generators. Section \ref{sec:Approach} describes the proposed B/F language models in detail. Section \ref{sec:Evaluation} presents experimental results. Finally, we have conclusion in Section \ref{sec:Conclusion}.

\section{Background and Related Work}\label{sec:Related}

\subsection{Language Modeling}
Given a corpus $\bm w=w_1, \cdots, w_m$, language modeling aims to minimize the joint distribution of $\bm w$, i.e. $p(\bm w)$. Inspired by the observation that people always say a sentence from the beginning to the end, we would like to decompose the joint probability as\footnote{
$w_1,w_2,\cdots,w_t$ is denoted as $\bm w_{1}^{t}$ for short.
} 
\begin{equation}
p(\bm w)=\prod_{t=1}^mp(w_t|\bm w_{1}^{t-1})\label{eqn:LM}
\end{equation}
Parameterizing by multinomial distributions, we need to further simplify the above equation in order to estimate the parameters. Imposing a Markov assumption---a word is only dependent on previous $n-1$ words and independent of its position---results in the classic $n$-gram model, where the joint probability is 
given by 
\begin{equation}
p(\bm w)\approx\prod_{t=1}^mp\left(w_t\big|\bm w_{t-n+1}^{t-1}\right)\label{eqn:Ngram}
\end{equation}
To mitigate the data sparsity problem, a variety of smoothing methods have been proposed. We refer interested readers to textbooks like \cite{NLP} for $n$-gram models and their variants.

Bengio et al.~\cite{NPLM}~propose to use feed-forward neural networks to estimate the probability in Equation~\ref{eqn:Ngram}. In their model, a word is first mapped to a small dimensional vector, known as an \textit{embedding}; then a feed-forward neural network propagates information to a softmax output layer, which estimates the probability of the next word.

A recurrent neural network (RNN) can also be used in language modeling. It keeps a hidden state vector ($\bm h_t$ at time $t$),  dependent on the its previous state ($\bm h_{t-1}$) and the current input vector $\bm x$, the word embedding of the current word.
An output layer estimates the probability that each word occurs at this time slot.
Following are listed the formulas for a vanilla RNN.\footnote{$W$'s refer to weights; biases are omitted.}
\begin{align}
\bm h_{t} &= \operatorname{RNN}(\bm x_t, \bm h_{t-1})\\
&=f\left(W_{\text{in}}\bm x_t+ W_{\text{hid}}\bm h_{t-1}\right)\\
p(w_t|\bm w_0^{t-1}) &\approx\operatorname{softmax}\left(W_\text{out}\bm h_{t}\right)
\end{align}
As is indicated from the equations, an RNN provides a means of direct parametrization of Equation~\ref{eqn:LM}, and hence has the ability to capture long term dependency, compared with $n$-gram models. In practice, the vanilla RNN is difficult to train due to the \textit{gradient vanishing or exploding} problem; long short term (LSTM) units \cite{LSTM} and gated recurrent units (GRU)~\cite{GRU} are proposed to better balance between the previous state and the current input.

\subsection{Language Generation}\label{ssec:Generation}
Using RNNs to model the joint probability of language makes it feasible to generate new sentences. An early attempt generates texts by a character-level RNN language model \cite{charRNN}; recently, RNN-based language generation has made breakthroughs in several real applications.

The sequence to sequence machine translation model \cite{seq2seq} uses an RNN to encode a source sentence (in foreign language) into one or a few fixed-size vectors; another RNN then decodes the vector(s) to the target sentence. Such network can be viewed as a language model, conditioned on the source sentence. At each time step, the RNN predicts the most likely word as the output; the embedding of the word is fed to the input layer at next step. The process continues until the RNN generates a special symbol $<$eos$>$ indicating the end of the sequence.
Beam search \cite{seq2seq} or sampling methods \cite{sigdial} can be used to improve the quality and diversity of generated texts.
 
If the source sentence is too long to fit into one or a few fixed-size vectors, an attention mechanism~\cite{attention} can be used to dynamically focus on different parts of the source sentence during target generation. In other studies, Wen et al.~use an RNN to generate a sentence based on some abstract representations of semantics; they feed a one-hot vector, as additional information, to the RNN's hidden layer \cite{sigdial}. In a question answering system, Yin et al.~leverage a soft logistic switcher to either generate a word from the vocabulary or copy the candidate answer \cite{QA}.

\section{The Proposed B/F Language Model}\label{sec:Approach}

In this part, we introduce our B/F language model in detail. Our intuition is to seek a new approach to decompose the joint probability of a sentence (Equation~\ref{eqn:LM}). If we know \textit{a priori} that a word $w_s$ should appear in the sentence ($\bm w = w_1, \cdots, w_m$, $w_s\in\bm w$), it is natural to design a Bayesian network where $w_s$ is the root node, and other words are conditioned on $w_s$. Following the spirit of ``sequence'' generation, $w_s$ split the sentence into two subsequences:
\begin{itemize}
\item Backward sequence: $w_s, w_{s-1}, w_{s-2},\cdots, w_{1}$\\
($s$ words)
\item Forward sequence: $w_s, w_{s+1}, w_{s+2},\cdots, w_{n}$\\
($m-s+1$ words)
\end{itemize}

The probability that the sentence $\bm w$ with the split word at position $s$ decomposes as follows.\footnote{$p({\operatorname*{=}\limits_\cdot^{\cdot}})$ denotes the probability of a particular backward/forward sequence.}

\begin{equation}
p\left(\begin{array}{l}
\bm w_s^1\\\hline\hline
\bm w_s^{n}
\end{array}\right)=p(w_s)\prod_{i=0}^{s}p^{\text{(bw)}}(w_{s-i}|\cdot)\prod_{i=0}^{m-s+1}p^{\text{(fw)}}(w_{s+1}|\cdot)\label{eqn:FB}
\end{equation}

\begin{figure}[!t]
\centering
\includegraphics[width=.6\textwidth]{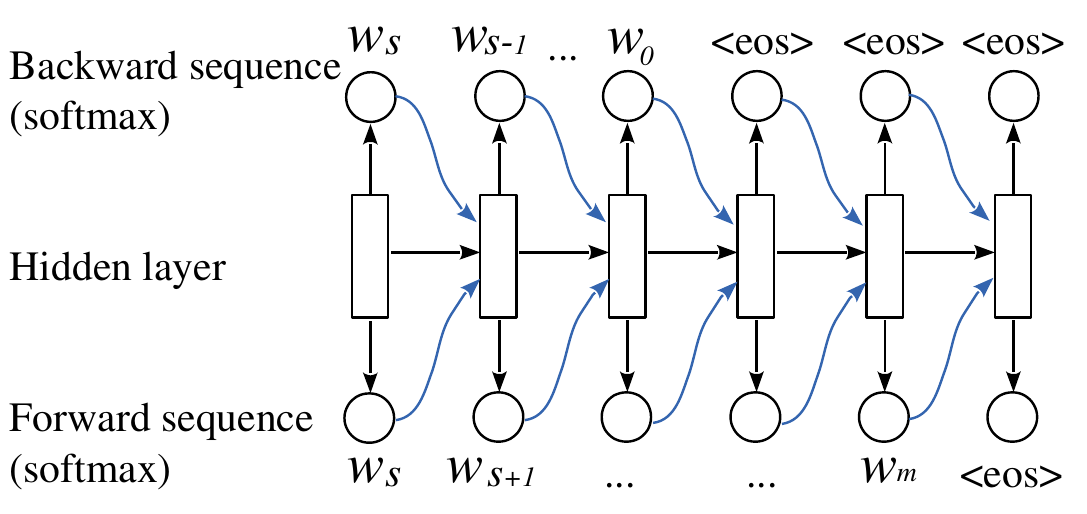}
\caption{Synchronous foward/backward generation.
}\label{fig:syn}
\end{figure}

To parametrize the equation, we propose two model variants. The first approach is to generate previous and backward models simultaneously, and we call this \textbf{syn-B/F} language model (Figure~\ref{fig:syn}).\footnote{Previously called backbone LM.} Concretely, Equation~\ref{eqn:FB} takes the form
\begin{equation}
p\left(\begin{array}{l}
\bm w_s^1\\\hline\hline
\bm w_s^{n}
\end{array}\right)=\prod_{t=0}^{\max\{s, m-s+1\}-1}
p\left(\begin{array}{l}
w_{s-t}\\\hline\hline
w_{s+t}
\end{array}\Bigg|\begin{array}{l}
\bm w_s^{s-t+1}\\\hline\hline
\bm w_s^{s+t-1}
\end{array}
\right)
\label{eqn:syn}
\end{equation}

\begin{figure}[!]
\centering
\includegraphics[width=.6\textwidth]{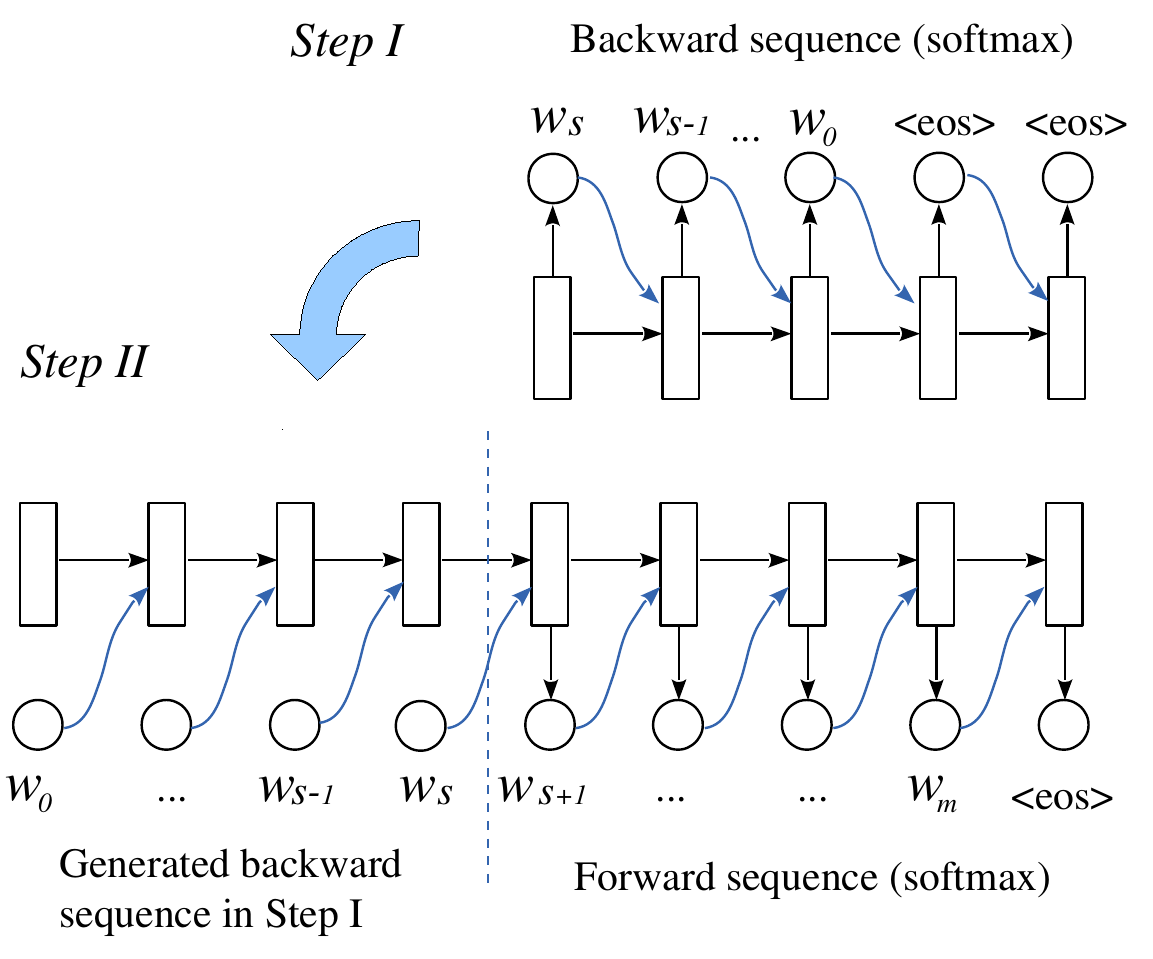}
\caption{Asynchronous forward/backward generation.}
\label{fig:asyn}
\end{figure}

\noindent where the factor $p(\text{=}|\text{=})$ refers to the conditional probability that current
time step $t$ generates $w_{s-t}, w_{s+t}$ in the forward and backward sequences, respectively, given the middle part of the sentence, that is, $w_{s-t+1}\cdots w_s\cdots w_{s+t-1}$. If one part has generated $<$eos$>$, we pad the special symbol $<$eos$>$ for this sequence until the other part also terminates.

Following the studies introduced in Section \ref{sec:Related}, we also leverage a recurrent neural network (RNN) . The factors in Equation~\ref{eqn:syn} are computed by
\begin{align}&p\left(\begin{array}{l}
w_{s-t}\\\hline\hline
w_{s+t}
\end{array}\Bigg|\begin{array}{l}
\bm w_s^{s-t+1}\\\hline\hline
\bm w_s^{s+t-1}
\end{array}
\right)\\
=& p^{\text{(bw)}}\left(w_{s-t}|\bm h_t\right)\cdot p^{\text{(fw)}}\left(w_{s+t}|\bm h_t\right)\\
=& \operatorname{softmax}\left(W_\text{out}^\text{(bw)}\bm h_t\right)\cdot
\operatorname{softmax}\left(W_\text{out}^\text{(fw)}\bm h_t\right)
\end{align}

Here, $\bm h_t$ is the hidden layer, which is dependent on the previous state $\bm h_{t-1}$ and current input word embeddings $\tilde{\bm x}=\left[\bm x_t^\text{(fw)}; \bm x_t^\text{(bw)}\right]$. We use GRU \cite{GRU} in our model, given by 
\begin{align}
\bm r &= \sigma(W_r\tilde{\bm x} + U_r\bm h_{t-1})\\
\bm z &= \sigma(W_z\tilde{\bm x} + U_z\bm h_{t-1})\\
\tilde{\bm h}&=\tanh\big(W_x\tilde{\bm x}+U_x(\bm r\circ\bm h_{t-1})\big)\\
\bm h_t&=(1-\bm z)\circ \bm h_{t-1}+\bm z\circ\tilde{\bm h}
\end{align}
where $\sigma(\cdot)=\frac{1}{1+e^{(-\cdot)}}$; $\circ$ denotes element-wise product.
$\bm r$ and $\bm z$ are known as gates, $\tilde{\bm h}$ the candidate hidden state at the current step.

In the syn-B/F model, we design a single RNN to generate both chains in hope that each is aware of the other's state. Besides, we also propose an asynchronous version, denoted as \textbf{asyn-B/F} (Figure~\ref{fig:asyn}). The idea is to first generate the backward sequence, and then feed the obtained result to another forward RNN to generate future words. The detailed formulas are not repeated.

It is important to notice that asyn-B/F's RNN for backward sequence generation is different from a generic backward LM. The latter is presumed to model a sentence from the last word to the first one, whereas our backward RNN is, in fact, a ``half'' LM, starting from $w_s$.

\bigskip
\noindent\textbf{Training Criteria.}
If we assume $w_s$ is always given, the training criterion shall be the cross-entropy loss of all words in both chains except $w_s$. We can alternatively penalize the split word $w_s$ in addition, which will make it possible to generate an entire sentence without $w_s$ being provided. We do not deem the two criteria differ significantly, and adopt the latter one in our experiments.

Both labeled and unlabeled datasets suffice to train the B/F language model. If a sentence is annotated with a specially interesting word $w_s$, it is natural to use it as the split word. For an unlabeled dataset, we can randomly choose a word as $w_s$.

Notice that Equation~\ref{eqn:FB} gives the joint probability of a sentence $\bm w$ with a particular split word $w_s$. To compute the probability of the sentence, we shall marginalize out different split words, i.e.,  
\begin{equation}
p(\bm w)=\sum_{s=1}^m\ p\left(\begin{array}{l}
\bm w_s^1\\\hline\hline
\bm w_s^{n}
\end{array}\right)
\label{eqn:backboneLM}
\end{equation}
In our scenarios, however, we always assume that $w_s$ is given in practice. Hence, different from language modeling in general, the joint probability of a sentence is not the number one concern in our model.

\bigskip

\section{Evaluation}\label{sec:Evaluation}
\subsection{The Dataset and Settings}
To evaluate our B/F LMs, we prefer a vertical domain corpus with interesting application nuggets instead of using generic texts like Wikipedia. In particular, we chose to build a language model upon scientific paper titles on arXiv.\footnote{http://arxiv.org
} Building a language model on paper titles may help researchers when they are preparing their drafts. Provided a topic (designated by a given word), constrained natural language generation could also acts as a way of brainstorming.\footnote{The title of this paper is NOT generated by our model.}

We crawled computer science-related paper titles from January  2014 to November 2015.\footnote{Crawled from http://http://dblp.uni-trier.de/db/journals/corr/} Each word was decapitalized, but no stemming was performed. Rare words ($\le10$ occurrences) were grouped as a single token, $<$unk$>$, (referring to \textit{unknown}). We removed non-English titles, and those with more than three $<$unk$>$'s. We notice that $<$unk$>$'s may appear frequently, but a large number of them refer to acronyms, and thus are mostly consistent in semantics.

Currently, we have 25,000 samples for training, 1455 for validation and another 1455 for testing; our vocabulary size is 3380. The asyn-B/F has one hidden layer with 100 units; syn-B/F has 200; This makes a fair comparison because syn-B/F should simultaneously learn implicit forward and backward LMs, which are completely different.
In our models, embeddings are 50 dimensional, initialized randomly. To train the model, we used standard backpropagation (batch size 50) with element-wise gradient clipping. Following \cite{visualize}, we applied rmsprop for optimization (embeddings excluded), which is more suitable for training RNNs than na\"ive stochastic gradient descent, and less sensitive to hyperparameters compared with momentum methods. Initial weights were uniformly sampled from $[-0.08, 0.08]$.  Initial learning rate was 0.002, with a multiplicative learning rate decay of 0.97, moving average decay 0.99, and a damping term $\epsilon=10^{-8}$.  As word embeddings are sparse in use \cite{regularization}, they were optimized asynchronously by pure stochastic gradient descent with learning rate being divided by $\sqrt{\epsilon}$.\footnote{The implementation was based on \cite{mou2,mou1}.}

\subsection{Results}
We first use the perplexity measure to evaluate the learned language models.
Perplexity is defined as $2^{-\ell}$, where $\ell$ is the log-likelihood (with base 2), averaged over each word.
$$\ell=\frac1m\sum_{i=1}^m\log p(w_i)$$
Note that $<$eos$>$ is not considered when we compute the perplexity.

We compare our models with several baselines:
\begin{itemize}
\item \textbf{Sequential LM}: A pure LM, which is not applicable to constrained sentence generation.
\item \textbf{Info-all}: Built upon sequential LM, {\tt Info-all} takes the wanted word's embedding as additional input at each time step during sequence generation, similar to \cite{sigdial}.
\item \textbf{Info-init}: The wanted word's embedding is only added at the first step (sequence to sequence model \cite{seq2seq}).
\item \textbf{Sep-B/F}: We train two separate forward and backward LMs (both starting from the split word).
\end{itemize}

\begin{figure}[!t]
\centering
\includegraphics[width=.45\textwidth]{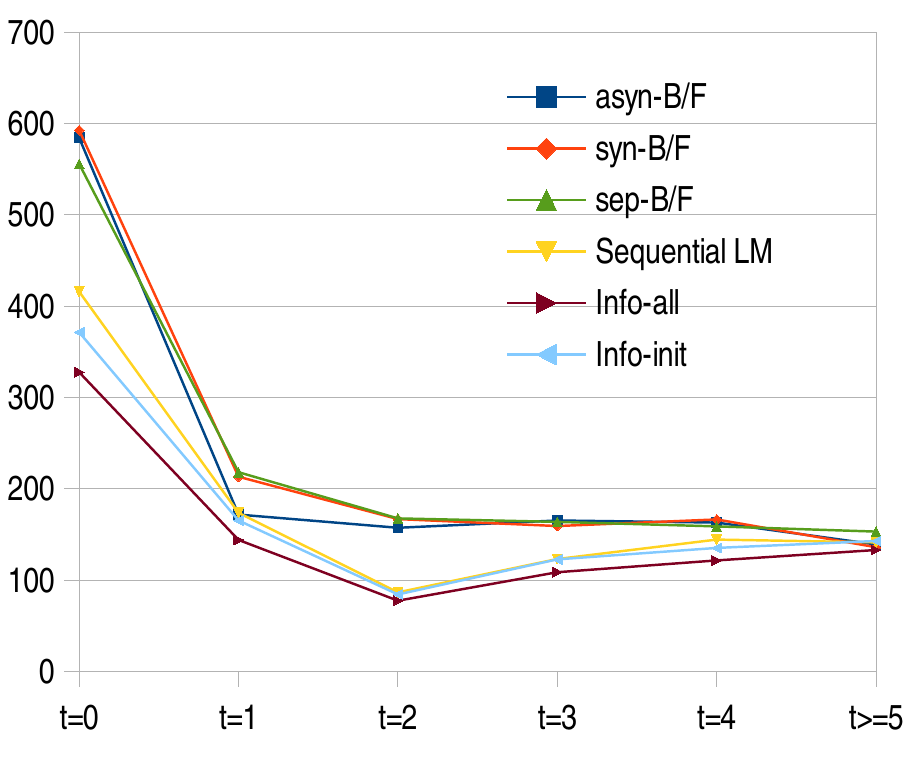}
\caption{Perplexity versus word position $t$, which is the distance between the current word and the first / split word in sequential, B/F LMs, respectively.
}\label{fig:PPL}
\end{figure}

Table~\ref{tab:PPL} summarizes the perplexity of our models and baselines. We further plot the perplexity of a word with respect to its position when generation (Figure~\ref{fig:PPL}).

\begin{table}[!t]
\centering
\resizebox{.8\textwidth}{!}{
\begin{tabular}{r|ccc}
\textbf{Method} & \textbf{Overall PPL} 
& \textbf{First word's PPL} 
& \textbf{Subsequent words' PPL}\\
\hline
Sequential LM & 152.2     & 416.2 & 134.8\\
Info-init & 148.7 & 371.5 & 133.3  \\
Info-all     &  125.4 & 328.0 & 121.8 \\
\hline
sep-B/F  & 192.4     & 556.1 & 169.9\\
sep-B/F ($w_s$ oracle) & 99.2 & -- & --\\
\hline
syn-B/F  & 185.4   & 592.7 & 162.9\\
syn-B/F ($w_s$ oracle) & 97.5 & -- & --\\
\hline
asyn-B/F & 177.2 & 584.5 & 153.7\\
asyn-B/F ($w_s$ oracle)& \textbf{89.8} & -- & --\\
\hline
\end{tabular}
}
\caption{Perplexity (PPL) of our B/F LMs and baselines.}
\label{tab:PPL}
\end{table}

\begin{table}[!t]
\resizebox{\textwidth}{!}{
\small
\begin{tabular}{p{.2\textwidth}|p{.2\textwidth}|p{.2\textwidth}|p{.2\textwidth}|p{.2\textwidth}}
\textbf{asyn-B/F} & \textbf{syn-B/F} & \textbf{Sep-B/F} & \textbf{Sequential LM}  & \textbf{Info-all}\\
\hline
deep \textbf{convolutional} neural networks for unk - based image segmentation&
 \textbf{convolutional} neural networks for unk - unk &
deep \textbf{convolutional} neural networks &
\textbf{convolutional} neural networks for unk -based object detection &
\textbf{convolutional} neural networks for image classification\\
\hline
object \textbf{tracking} and unk for visual recognition & 
learning deep convolutional features for object \textbf{tracking} &
object \textbf{tracking} &
\textbf{tracking} - unk - based  social media &
unk - based unk detection for image segmentation\\
\hline
optimal control for unk \textbf{systems} with unk - type ii : a unk - unk approach & 
formal verification of unk - unk \textbf{systems} &
optimal control for unk \textbf{systems} &
\textbf{systems} - based synthesis for unk based diagnose &
a new approach for the unk of the unk - free problem\\
\hline
unk \textbf{:} a new method for unk based on - line counting on unk
 & unk \textbf{:} a new approach for unk - based unk&
unk \textbf{:} a survey&
\textbf{:} a unk - based approach to unk - based deign of unk -based image retrieval &
unk \textbf{:} a unk - based approach for unk - free grammar\\
\hline
an approach \textbf{to} unk the edge -  preserving problem 
& an approach \textbf{to} unk &
an approach \textbf{to} unk - unk&
\textbf{to} unk : a unk - efficient and scalable framework for the unk of unk&
a unk - based approach \textbf{to} unk for unk\\
\hline
\end{tabular}
}

\caption{Generated texts by the B/F and sequential LMs, with the word in bold being provided.}
\label{tab:case}
\end{table}

From the experimental results, we have the following observations.
\begin{itemize}
\item All B/F variants yield a larger perplexity than a sequantial LM. This makes much sense because randomly choosing a split word increases uncertainly. It should be also noticed that, in our model, the perplexity reflects the probability of a sentence with a specific split word, whereas the perplexity of the sequential LM assesses the probability of a sentence itself.
\item Randomly choosing a split word cannot make use of position information in sentences. The titles of scientific papers, for example, oftentimes follow templates, which may begin with ``\textit{$<$unk$>$ : an approach}'' or ``\textit{$<$unk$>$ - based approach}.'' Therefore, sequential LM yields low perplexity when generating the word at a particular position ($t=2$), but such information is smoothed out in our B/F LMs because the split word is chosen randomly.
\item When $t$ is large (e.g., $t\ge4$), B/F models yield almost the same perplexity as sequential LM. The long term behavior is similar to sequential LM, if we rule out the impact of choosing random words. For syn-B/F, in particular, the result indicates that feeding two words' embeddings  to the hidden layer does not add to confusion.
\item In our applications, $w_s$ is always given, which indicates $p(w_s)=1$ (denoted as $w_s$ \textit{oracle} in Table~\ref{tab:PPL}). This reduces the perplexity to less than 100, showing that our B/F LMs can well make use of such information that some word should appear in the generated text. Further, our syn-B/F is better than na\"ive sep-B/F; asyn-B/F is further capable of integrating information in backward and forward sequences.
\end{itemize}

We then generate new paper titles from the learned language model with a specific word being given, which can be thought of, in the application,  as a particular interest of research topics. Table~\ref{tab:case} illustrates examples generated from B/F models and baselines. As we see, for words that are common at the beginning of a paper title---like the adjective \textit{convolutional} and gerund \textit{tracking}---sequential LM can generate reasonable results. For plural nouns like \textit{systems} and \textit{models}, the titles generated by sequential LM are somewhat influent, but they basically comply with grammar rules. For words that are unlikely to be the initial word, sequential LM fails to generate grammatically correct sentences.

Adding additional information does guide the network to generate sentences relevant to the topic, but the wanted word may not appear. The problem is also addressed in \cite{sigdial}.

By contrast, B/F LMs have the ability to generate correct sentences. But the sep-B/F model is too greedy in its each chain. As generating short and general texts is a known issue with neural network-based LMs, sep-B/F can hardly generate a sentence containing much substance. syn-B/F is better, and asyn-B/F is able to generate sentences whose quality is comparable with sequential LMs.

\section{Conclusion}\label{sec:Conclusion}
In this paper, we proposed a backward and forward language model (B/F LM) for constrained natural language generation. Given a particular word, our model can generate previous words and future words either synchronously or asynchronously. Experiments show a similar perplexity to sequential LM, if we disregard the perplexity introduced by random splitting. Our case study demonstrates that the asynchronous B/F LM can generate sentences that contain the given word and are comparable to sequential LM in quality.

\bibliographystyle{abbrv}
\bibliography{LM}
\end{CJK}
\end{document}